# Modified Apriori Graph Algorithm for Frequent Pattern Mining

Pritish Yuvraj
Dept of CSE
Bangalore Institute of Technology
Bangalore, India.
*pritish.yuvraj@yahoo.in*

Suneetha K. R.
Assisant Professor, Dept. of CSE
Bangalore Institute of Technology
Bangalore, India.
*suneetha.bit@gmail.com*

*Abstract*— Web Usage Mining is an application of Data Mining Techniques to discover interesting usage patterns from web data in order to understand and better serve the needs of web based applications. The paper proposes an algorithm for finding these usage patterns using modified version of Apriori Algorithm called Apriori-Graph. These rules will help service providers to predict, which web pages, the user is likely to visit next. This will optimize the website in terms of efficiency, bandwidth and will have positive economic benefits for them. The proposed Apriori Graph Algorithm $O((V)(E))$ works faster compared to the existing Apriori Algorithm and is well suitable for real time application.

## I. INTRODUCTION

Data Mining is the process of analyzing data from different perspectives and summarizing it into useful information that can be used to increase revenue, cut costs or both. Web Mining is the application of data mining techniques to discover patterns from the World Wide Web. It can be divided into three different types – Web usage mining, Web content mining and Web structure mining. Web usage mining itself can be classified further depending on the kind of usage data considered: Web Server Data, Application Server Data, and Application Level Data. Web log Mining includes three main stages: Data Pre-Processing, Pattern Discovery and Pattern Analysis.

A) Data Pre-Processing:

Web Server Data contains information such as who accessed the web site, what pages were accessed, Time of request etc. In pre-processing [3] stage, irrelevant data fields are removed and unique users are identified [4]. Transaction table is created through the user sessions. [5] [6] [17].

B) Pattern Discovery:

To discover patterns several data mining algorithms are available such as Path Analysis, Association Rules, Sequential Pattern, and Clustering [7].

C) Pattern Analysis:

Service Providers are particularly interested in knowing where the next click of mouse button will be by the user surfing their website. It helps them remain competitive, decreases bandwidth and improves overall efficiency. After analyzing the rules generated by pattern discovery algorithms website designers might restructure their websites accordingly. In recent years, there has been much research on Web Usage Mining [18], [19], [20], [21].

One of the popular algorithm for frequent set generation is Apriori [1] proposed by Rakesh Agarwal and Ramakrishnan Srikant. The algorithm $2^N$ computations for pattern generation. If 100 frequent elements were found after a database scan, Apriori will make $2^{100}$ computations. Second drawback is it's time complexity $O(e^N)$ [2]. Third, the number of database scans by the Apriori Algorithm is huge, hence incurring more I/O load. These features of Apriori make it undesirable for frequent set generation, hence modification of Apriori Algorithm is necessary in order to reduce execution time. A lot of research has been conducted to improve Apriori Algorithm.

Towards this direction a modified version of Apriori algorithm is proposed in this paper Apriori Graph to generate frequent patterns. Concept of graphs and Apriori are collectively used. This enables mining of the graph for finding patterns rather than going back to the database at every pass. This limits the number of Input/Output operations to two (major disadvantage of Apriori), also the number of total combinations is brought down from $2^N$ to the total number of edges E formed by the graph. This considerably brings down the time and makes real time pattern discovery possible. The proposed algorithm is of $O((V)(E))$. This is discussed in section III in details.

Layout of the paper is: Section II discusses Related Works, III contains the proposed algorithm along with its description, IV provides Architecture, V presents Results and Analysis, VI contains Conclusion, followed by References.

## II. RELATED WORKS

A Web log file [12] records activity information when a Web user submits a request to a Web Server. The ability of using the data mining techniques to extract information from the server logs was first introduced by [13], [14], and [15].

First step in any web usage mining task is preprocessing. K. R. Suneetha and Dr. R. Krishnamoorthi [3] give an insight upon how important it is to properly preprocess the log data. Data Cleaning is done to remove the inappropriate records with unsuccessful status [16].

Agarwal et all [8] presented AIS algorithm which generates candidate item sets during each pass of the database scan. This algorithm turns out to be effete as it generates too many candidate item sets. The algorithm also requires much space and time. Also due to too many passes over the database results in increase I/O load.

Apriori algorithm [1] proposed by Rakesh Agrawal and Ramakrishnan Srikant is an influential data mining algorithm. It is an algorithm which can solve the problem of web usage mining. It generates a list of most frequent web pages visited. Being a very slow algorithm of the order $O(E^N)$ [2] is the biggest disadvantage for the service providers. Due to fast changing contents of database one needs an algorithm which is real time.

Direct Hashing and Pruning (DHP) [10] algorithm was developed to reduce size of candidate set by filtering any k-item sets out of the hash table. This powerful filtering capability allows DHP to complete execution when Apriori is still at its second pass and hence shown improvement in execution time and utilization of space.

Another way to improve Apriori is to use most suitable data structure such as frequent pattern tree. Han et. al., in [22] introduced an algorithm known as FP-Tree algorithm for frequent pattern mining. It is another milestone in the development of association rule mining and avoids the candidate generation process with less passes over the database. FP-Tree algorithm breaks the bottlenecks of Apriori series algorithms but suffers with limitations. It is difficult to use in an environment that users may change the support threshold with regard to the mining results, and once the support threshold changed, the old FP-Tree cannot be used anymore, hence additional effort is needed to re-construct the corresponding FP-Tree. It is not suitable for incremental mining, since as time goes on databases keep changing, new datasets may be inserted into the database or old datasets be deleted, and hence these changes lead to a re-construction of the FP-Tree[6].

Data Mining includes another important issue of Scalability. Algorithms must be able to "Scale up" to handle large amount of data. Eui-Hong et. al [11] tried to make data distribution and candidate distribution scalable by Intelligent Data Distribution (IDD) algorithm and Hybrid Distribution (HD) algorithm respectively. IDD addresses the issues of communication overhead and redundant computation by using aggregate memory to partition candidates and move data efficiently. HD improves over IDD by dynamically partitioning the candidate set to maintain good load balance.

The quality of association rule discovered is measured in terms of confidence. The rules with confidence above a threshold are considered. Most of the Algorithms ask this threshold from the user whereas APACS2 do not. APACS2 [9] is an algorithm which make use of an objective interestingness measure called adjusted differences. It discovers positive and negative association rules. APACS2 uses adjusted difference as an objective interestingness' measure. Adjusted differences are defined in terms of Laws of Standardized differences and maximum likelihood estimate.

An improved version of original Apriori- All algorithm is developed for sequence mining in [15]. It adds the property of the userID during every step of generation of candidate set and every step of scanning the database to decide about whether an item in the candidate set should be used to produce next candidate set. The algorithm reduces the size of candidate set in order to reduce the number of database scanning.

To overcome this difficulty modified version of Apriori Algorithm is proposed to generate frequent web pages. After first pass, list of all frequent web pages are determined. After second pass a correlation is found between frequent web pages from the database. Now these correlations are put in form of a graph. The graph is mined for finding patterns instead of the database. Detailed information is provided in section III.

III. PROPOSED METHODOLOGY & DESCRIPTION

Psuedo Code for the Proposed Apriori with Graph Algorithm:

```
1) L1 = { large 1 itemsets}
2) webaddr = [][]
3) foreach item1 in L1:
4)     foreach item1 in L1:
5)         webaddr[item1][item1] = 0
6) foreach transaction t in D:
7)     ct = subset_of_t_of_length_2(t):
8)     forall candidates ct in webaddr:
9)         webaddr[ct1][ct2] += 1
10) Answer = []    //Global Variable
11) result_list = []
12) foreach vertex1 in L1:
13)     empty result_list
14)     result_list.append(vertex1)
15)     AprioriGraph(result_list, vertex1, webaddr)
16) Rules = [] //global variable
17) generate_rules(Answer)
```

Fig. 1 Pseudo-Code for the Proposed Apriori Graph

*AprioriGraph*(result_list, vertex1, webaddr):
```
1) count = 0
2) foreach vertex2 in webaddr[vertex1]:
3)     if((webaddr[vertex1][vertex2]>Confidence) && (vertex2 not in result_list)):
4)         result_list.append(vertex2)
5)         count ++
6)         AprioriG(result_list, vertex2, webaddr)
7) if (count<1) and (not all elements of result_list present in any list of Answer):
8)     found_conf = 0
9)     forall t in D:
10)        forall elements of result_list in t:
11)            found_conf ++
12)    if found_conf >= confidence:
13)        Answer.append([result_list])
14) delete last element from result_list
15) return
```

Fig. 2 Pseudo Code for Mining graph

*generate_rules*(Answers):
1) foreach result in Answer:

```
2)      ct = all_subsets(result)
3)      foreach item in ct:
4)          remaining = result - item
5)          if len(remaining)>0:
6)              support = getsupport(result)/getsupport(item)
7)              if support > req_support:
8)                  Rules.append([[item],[remaining],[confidence]])
```
Fig. 3 Pseudo Code for Generation of Rules

The following example in Fig, 4 will explain the proposed Algorithm:

Confidence is set to 60% and Support to 20%. No of transactions are 9, so support no is 0.2*9 = 1.8 or 2. Frequent Pattern and rules will be generated after the execution of the Proposed Algorithm.

| tiD | List of item's in each transaction |
|---|---|
| 1 | Milk, Butter, Sugar |
| 2 | Butter, Beer |
| 3 | Butter, Bread |
| 4 | Milk, Butter, Beer |
| 5 | Milk, Bread |
| 6 | Butter, Bread |
| 7 | Milk, Bread |
| 8 | Milk, Butter, Bread, Sugar |
| 9 | Milk, Butter, Bread |

Fig. 4 Sample Database

(Fig.1 - Line1) Finds all the items which occur more than 1.8 times or precisely atleast 2 times from the sample database (Fig. 4). The items which qualify are Milk( count: 6), Butter( count: 7 ), Bread( count: 6 ), Beer( count: 2 ), Sugar( count: 2 ).

(Fig.1 - Line3-5) Initializes a 2D matrix called *webaddr* as shown in Fig. 5 with zeros. Assume for a 2D matrix, vertical coordinates of the graph are named as the X coordinate and horizontal coordinates are called Y coordinate. The X and Y coordinates of the matrix are names of the items which qualified after first step. i.e. Milk, Butter, Bread, Beer, Sugar as shown in tht fig. below.

|  | Milk | Butter | Bread | Beer | Sugar |
|---|---|---|---|---|---|
| Milk | 0 | 0 | 0 | 0 | 0 |
| Butter | 0 | 0 | 0 | 0 | 0 |
| Bread | 0 | 0 | 0 | 0 | 0 |
| Beer | 0 | 0 | 0 | 0 | 0 |
| Sugar | 0 | 0 | 0 | 0 | 0 |

Fig. 5 matrix *webaddr* initialized with 0's

(Fig.1 Line 6) a loop is run for every transaction in the sample database (Fig. 4) from tiD = 1 to tiD = 9.

(Fig. 1 Line 7) generates combinations of two items for every transaction (Fig. 4) . Example, in Fig. 4 for tiD = 1, items are (Milk, Butter, Sugar). Combinations of two items are (Milk, Butter), (Butter, Sugar), (Milk, Sugar).

(Fig.1 Line 8-9) Checks whether the combinations generated *eg. [ [Milk. Butter] (Milk as X-Coordinate and Butter as Y-Coordinate) ]* are present in matrix *webaddr* (Fig. 5). If the combinations are present then increase the value by 1. Eg. (Milk, Butter) is present in *webaddr* (Fig. 5) so increase the value by 1.

After the execution of all the transactions in sample database (Fig. 4), matrix *webaddr* has weights as shown in Fig. 6

|  | Milk | Butter | Bread | Beer | Sugar |
|---|---|---|---|---|---|
| Milk | 0 | 4 | 4 | 1 | 2 |
| Butter | 0 | 0 | 3 | 1 | 2 |
| Bread | 0 | 0 | 0 | 0 | 1 |
| Beer | 0 | 0 | 0 | 0 | 0 |
| Sugar | 0 | 0 | 0 | 0 | 0 |

Fig. 6 Matrix *webaddr*

Fig. 6, shows that [Milk, Butter] as a combination occurred four times in Sample Database (Fig. 4), [Butter, Bread] as a combination occurred thrice in the sample database (Fig. 4). Counts in *webaddr* (Fig. 6) will now be referred as "edge weights". So, the edge weight between vertices [Milk, Butter] is four. The matrix represented in Fig. 6 is an efficient way to represent database in (Fig. 4). It took just two passes of database (Fig. 4) to structure the matrix in (Fig. 6). Instead of looking for patterns from the sample database (Fig. 4), matrix in (Fig. 6) will be mined to generate the same (Fig. 6).

(Fig. 1, Line 10) Global 2D list variable *"Answer"* is declared so that all discovered frequent patterns in the form of lists are stored in this variable. (Fig 1, Line 11) another list variable *"result_list"* is declared.

(Fig. 1, Line 12-15) Finds the frequent patterns from the graph in (Fig. 6). (Fig.1 Line 12), a loop is run for each vertex found in (Fig.1 Line 1). *i.e. [Milk. Butter, Bread. Beer, Sugar]*. (Fig. 1, Line 13) Every time *"for"* loop iterates, variable *"result_list"* is emptied. (Fig. 1, Line 14) vertex to be appended to the list in first iteration is *"Milk"*. (Fig. 1, Line 15) Recursive Function *AprioriGraph* (Fig. 2) is called with **parameters** *(result_list (value = [Milk]), vertex1 (value = Milk) and webaddr)* (Fig. 6).

*AprioriGraph* is a recursive algorithm which adds one new vertex from the graph *webaddr* (Fig. 6) at every recursive call if all the following three conditions are fulfilled:

1) The new vertex found must not be present in the list *result_list*.

2) The Edge weight from the last element in *result_list* in (Fig. 2) to the new vertex found must be >= Support Value. The edge weight can be found from *webaddr* in (Fig. 6)

3) Atleast 1 new vertex must be discovered fulfilling the above two criteria at every *AprioriGraph* function call. If not return back.

Initially list variable *"result_list"* is declared empty (Fig. 1 Line 12).

| Null | Null | Null | Null |

**Fig. I) result_list**

At the time of call to function *AprioriGraph* (Fig. 2) list type variable *"result_list"* has *"Milk"* appended to it (appended in Fig. 2 Line 13).

| **Milk** | Null | Null | Null |

**Fig. II) result_list**

(Fig. 2 Line 1) Declares a variable called *count*. This is used to keep a track whether any new vertex has been added during the present call of the recursive function. If not even a single new vertex fulfilling the three conditions previously mentioned are encountered *count* will have value as zero.

(Fig. 2 Line 2) vertex2 holds each one of the elements from y-coordinates of the graph *webaddr* (Fig. 6) with the x-coordinate as "Milk". Let in first instance vertex2 (from y-coordinate) hold Butter. i.e. *webaddr[Milk][Butter]*.

Edge weight from Milk → Butter is four. This is greater than the support value 2. Also, *result_list* contains [Milk] (fig. II), and there is no Butter in it. Hence Butter is accepted and stored in *result_list*.

| **Milk** | **Butter** | Null | Null |

**Fig. III) result_list**

Now recursive function *AprioriGraph* (Fig. 2 Line 6) is again called. This time for the function, *"vertex1"* which earlier had *"Milk"* will now have *"Butter"* (Fig. 2 Line 6). Also, (Fig. 2 Line 5) count is increased by one. This is important as redundancy needs to be removed. Eg. [A, B, C] is a frequent pattern, mentioning [A, B] again in frequent pattern is redundancy. Since we have increased the value of *count* by one, (Fig. 2 Line 7) will keep track of the fact that after Milk → Butter → some other vertex will also be visited. Hence higher recursive calls will save or print the result, present executing function need not bother. This prevents redundancy.

Coming back to the call to *AprioriGraph* with vertex1 now has Butter, the graph *webaddr* (Fig. 6) is again iterated. Butter → Bread is found with edge weight three >= support ( i.e. 2). Also, Bread is not present in *result_list* (fig. III). Again recursive call takes place and with vertex1 now as Bread.

| **Milk** | **Butter** | **Bread** | Null |

**Fig. IV) result_list**

A new vertex is searched from ( Bread → something ) that can satisfy the support value and is unique to *webaddr* but no such vertex is found. *"Count"* (Fig. 2) for this level of the recursive function remains zero. i.e. No new vertex added. Hence the elements in the *result_list* might be a frequent pattern.

(Fig. 2 Line 8-13) Checks the support for all three frequent patterns together (Milk → Butter → Bread) the database (Fig. 4) . This is necessary to break Transitivity. A → B and B → C implies A → C. But this rule might not hold when it comes to web usage mining, so this needs to be broken.

(Milk → Butter → Bread) appears in 2 tiD's (tid 8, tid 9 in Fig. 4) which is just in accordance with the support value. Hence it is accepted as a frequent pattern.

Line 13, deletes Bread (the last element) from the list and start scanning again as ( Butter → something).

| **Milk** | **Butter** | Null | Null |

**Fig. V) result_list**

It now finds (Butter → Beer) qualifying as count is 2. So the count for the new frequent pattern (Milk → Butter → Beer) is calculated from the sample database Fig. 4.

The above combination appears just once in the database which is < support value. i.e. 2 . Hence this is rejected as a frequent pattern. The pattern became frequent because of transitivity. (*Milk → Butter*) was a high frequent pattern and so was *(Butter → Beer)* but *(Milk → Butter → Beer)* does not hold true. This is how transitivity is broken by the proposed algorithm, an inevitable case in graph mining. Lastly two high frequency items found are:

1) [Milk, Butter, Sugar]
2) [Milk, Butter, Bread]

From the fig. 7, the frequent patterns derived are **a)** *[Milk, Bread]*, **b)** *[Milk, Sugar]*, **c)** *[Mik, Butter, Bread]*, **d)** *[Milk, Butter, Bread]*, **e)** *[Milk, Butter, Sugar]* **f)** *[Butter, Bread]* **g)** *[Butter, Sugar]*. But in the final high frequent list only **d)** and **e)** appears. The reason being (Fig. 2, Line 7) second condition in *if* statement. This statement takes care of the fact that if a new frequent pattern list is found (eg. *[Milk, Sugar])* and that all the items of the list are already present in some other appended list of variable *"Answer"* (i.e. *[Milk, Butter, Sugar]*, then the new frequent pattern found will be discarded on the grounds of redundancy.

Similarly, when all the edges originating with starting vertex "Milk" has been exhausted, "Butter" the immediate next x-coordinate to "Milk" will discover frequent patterns [*Butter, Bread]* and [*Butter, Sugar]* (Fig. 6) as they have sufficient edge weights but both of them will be discarded as they have already appeared in **d)** and **e)** of the previous paragraph. Hence the collection of the frequent patterns will be free from redundancy.

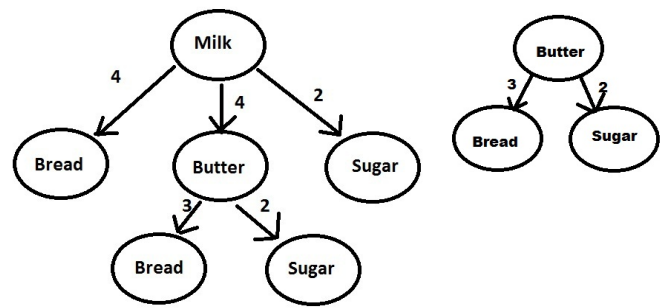

Fig. 7 (Tree structure formed by the proposed Apriori Graph Algorithm)

To generate rules from the frequent pattern the below formula is used:

Confidence (X → Y) = support(X U Y) / support(X) **(Eqn. 1)**

This can be understood as the Probability of occurrence of A, given B has already occurred.

P(A | B) = P( A ∩ B) / P (B)

Has a customer bought Milk and Sugar then  P(Butter | Milk ∩ Sugar) represents the probability that he will also purchase Butter.

(Fig. 3 ) *generate_rules* accomplishes rules generation from line 1-8.  List *"Answer"* is passed to the above function containing 2D list **[[Milk, Butter, Sugar], [Milk, Butter, Bread]].**

First frequent pattern is stored in list *"result" (*Fig. 3 Line. 1)i.e. [Milk, Butter, Sugar]. (Fig. 3 Line 2) generates all the subsets of the same. i.e. [Milk, Sugar], [Milk, Butter], [Milk], etc. and stores in a variable "c*t"*. A list is selected from "c*t"* and stored in "*item"* eg.[*Milk, Sugar]*.  List "*remaining"* (Fig.3 Line 4) stores the variables which were present in list "*result*" but not present in "*item"* i.e. *[Butter]*. Confidence is calculated as per **Eqn.1** and implemented in (Fig. 3, Line 6). All those rules which qualify are appended in the list *Rules*  in Fig. 3 Line 8.

For calculating confidence for (Milk, Sugar) → Butter. Milk, Butter and Sugar occur together in 2 transaction Id's. Milk and Sugar together occur in 2 transactions Id's.

Therefore, confidence = 2/2 = 1 or 100%

This means that, if a customer buys Milk and Sugar then 100% he will go for Butter.

These rules are very useful in Web Usage Mining. Web Servers can predict what the next click of the user can be and restructure themselves accordingly.

List of all the rules found for (Milk → Butter → Sugar) along with their confidence are:

1) (Milk, Sugar) → Butter     (100%)
2) (Butter, Sugar) → Milk     (100%)
3) (Sugar) → (Butter, Milk)     (100%)

A) Time Complexity:

For initialization of the 2D graph with 0's and for finding the counts of the pair of vertices it takes O(n2) time as both have 2 for loops each.

*AprioriGraph* is called for each vertex i.e. O(V). The total number of times the *AprioriGraph* function will recurse plus the for loops will run is <= total number of edges O(E) of *webaddr* (Fig. 7). So, *AprioriGraph*  is O((V)(E)) algorithm. Fig. 3, *generate_rules*  has two for loops each and the time complexity associated will be O(n2).

So the proposed Apriori-Graph algorithm is an O((V)(E)) algorithm.

IV. ARCHITECTURE

The architecture of the proposed Algorithm is shown in Fig. 8.

Fig. 8 Architecture of the Proposed Algorithm

Initially the raw database is collected from Server Logs. Clearing, Preprocessing, Identification of user and session is done from the database. The database hence generated is stored in a table format. The proposed *AprioriGraph* Algorithm(Fig. 1 & Fig. 2) is applied to the preprocessed data. Frequent patterns are generated having support value above a certain user threshold. From these frequent patterns, rules are generated (Fig. 3) which are above certain user mentioned confidence. Then these rules are provided to the service providers. Later the website designers can restructure the WebPages accordingly.

V. RESULT AND ANALYSIS

The NASA Web Access Service Log File (1995) database is used for comparing time, space and accuracy between the proposed algorithm and existing Apriori algorithm.

The system on which it was tested had 4GB RAM, i3 Processor and Ubuntu 14.04 (Linux) as the operating system. The codes were implemented in Python version 2.7. The original NASA web log file was of 205MB. It had columns where there were host, time-stamp, request, HTTP reply code and bytes in the reply as the fields.

Data was preprocessed. Host, requests were taken into consideration only if HTTP reply code was 200 (for success). Sessions of 30 mins were created. For every individual user, 'host' was considered as Transaction No and all the web pages they requested within the session of 30 mins were considered as items. Requests having '.gif' or icons like '.xbm' were cleaned as nobody would like to know after all the processing that they discovered the most frequent request from the user was for a web logo (This is quite possible because on every web page logo of the company is present as an image. Every time user loads a web page, icons automatically reload. Chances are very high that these logos will come in as most frequent requests by users).  Also eliminate refresh button hit by users from database. After preprocessing, the size of the database fell to 10MB.

As a part of the user input,  support was set to 5% and confidence to 50%. The results for both the algorithms are as follows. The rules generated by Apriori Algorithm are as follows:

| X | Y | (X → Y) Confidence |
|---|---|---|
| /shuttle/missions /sts-71/images /images.html | /shuttle /countdown/ | 57.1% |
| Shuttle/missions/sts-71/mission-sts-71.html | /shuttle /countdown/ | 58.1% |

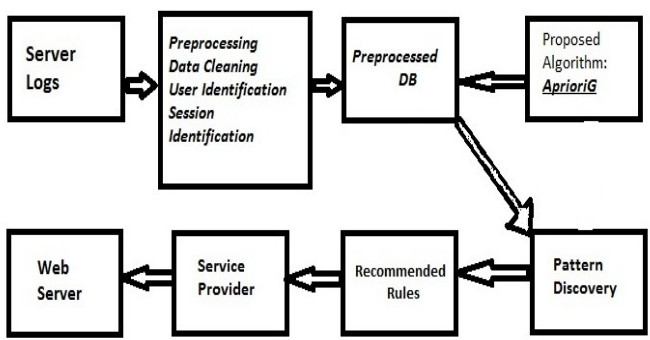

The following rules were generated by proposed Apriori Graph Algorithm.

| X | Y | (X → Y) Confidence |
|---|---|---|
| /shuttle/missions/sts-71/images/images.html | /shuttle/countdown/ | 57.1% |
| Shuttle/missions/sts-71/mission-sts-71.html | /shuttle/countdown/ | 58.1% |
| /shuttle/countdown/ | /shuttle/countdown/liftoff.html | 50.2% |

A comparison of space occupied in memory (MB) Vs amount of Data (MB) is shown below. '$top' command in Linux was used for the task. This command in Linux operating system shows the percentage of total memory consumed by a process. The X axis is space consumed in MB vs. Y axis is size of the data (MB).

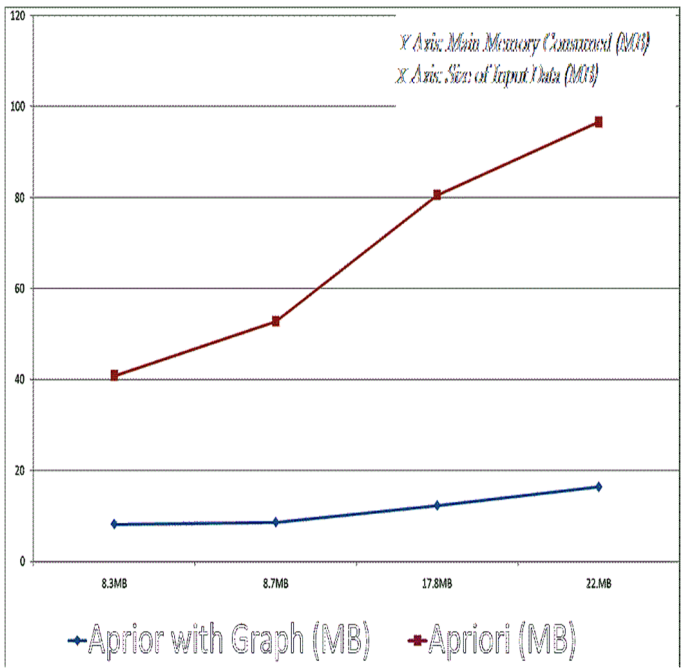

Fig. 9 (Comparison of Space consumed by Apriori Algorihm Vs proposed Apriori Algorithm)

Red line shows the space consumption by Apriori Algorithm whereas blue line shows space consumption by the proposed AprioriGraph Algorithm.The proposed Algorithm consumes considerably less memory compared to Apriori.
Comparison graph is drawn between the performances of Apriori with the proposed Algorithm below. On the X axis is time taken to find the rules (in secs) and on the Y axis is the amount of data given as input (MB).

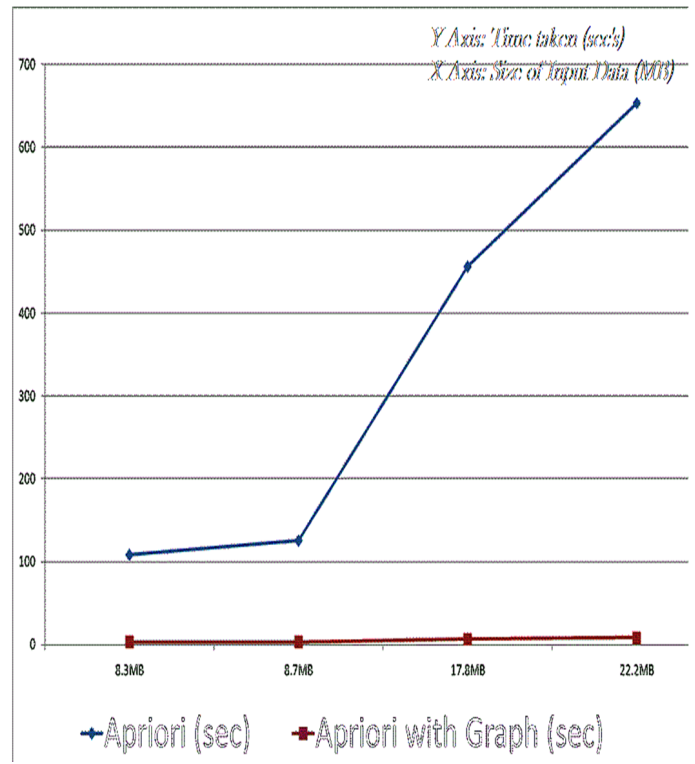

Fig 10. Comparison of time consumed by Apriori Algorithm Vs the proposed Apriori graph Algorithm.

Blue line shows the time consumed by Apriori Algorithm, whereas the red line shows time consumed by the proposed Apriori Algorithm. It is certain from the graph that the proposed Algorithm consumes much lesser time compared to Apriori Algorithm. Curve of the proposed Algorithm shows that its slope does not increase as quick as Apriori, this suggests that even if the data increases twice or thrice, performance will not suffer as considerably as Apriori.

VI. CONCLUSION

The proposed algorithm overcomes the deficiency present in existing Apriori Algorithm. The basic algorithm of Apriori scans the database many times incurring heavy Input/Output load. It generates huge number of combinations for finding frequent patterns. Also Apriori is an exponential time algorithm. All the above factors make it unviable for real time applications. The proposed algorithm is fast, efficient, takes less memory and implementable in places where real time associations are required.